\documentclass{article}

\usepackage[preprint]{wenxiaobai}

\usepackage[utf8]{inputenc} % allow utf-8 input
\usepackage[T1]{fontenc}    % use 8-bit T1 fonts
\usepackage{hyperref}       % hyperlinks
\usepackage{url}            % simple URL typesetting
\usepackage{booktabs}       % professional-quality tables
\usepackage{amsfonts}       % blackboard math symbols
\usepackage{nicefrac}       % compact symbols for 1/2, etc.
\usepackage{microtype}      % microtypography
\usepackage{xcolor}         % colors
\usepackage{atbegshi}
\usepackage{caption}

\usepackage{times}
\usepackage{latexsym}
\usepackage{listings}
\usepackage{wrapfig}
\usepackage[symbol]{footmisc}
\usepackage{mathptmx}
\usepackage{microtype}
\usepackage{tcolorbox}
\usepackage{tikz}

\usepackage{inconsolata}

\usepackage{graphicx}
\usepackage{multirow}
\usepackage{bbding}
\newcommand{\RealQuestionSmallCaps}{\textsc{RealQuestions}}
\newcommand{\SynthQuestionSmallCaps}{\textsc{SynthQuestions}}
\tcbuselibrary{breakable, skins}

\title{From \textsc{Real} to \textsc{Synthetic}: Synthesizing Millions of Diversified and Complicated User Instructions with Attributed Grounding}

\author{{\bf Chiwei Zhu\textsuperscript{\rm 1,2\S}, Benfeng Xu\textsuperscript{\rm 1,2\textdagger}, Xiaorui Wang\textsuperscript{\rm 2}, Zhendong Mao\textsuperscript{\rm 1}} \\
\textsuperscript{1}University of Science and Technology of China \\ 
\textsuperscript{2}Metastone Technology \\ 
\texttt{\{tanz, benfeng\}@mail.ustc.edu.cn}}

\begin{document}

\maketitle

\begin{abstract}
The pursuit of diverse, complex, and large-scale instruction data is crucial for automatically aligning large language models (LLMs). While there are methods capable of generating synthetic instructions at scale, they either suffer from limited grounding sources, leading to a narrow distribution, or rely on trivial extensions that fail to produce meaningful trajectories in terms of complexity. In contrast, instructions that benefit efficient alignment are typically crafted with cognitive insights and grounded in real-world use cases. In this paper, we synthesize such instructions using attributed grounding, which involves 1) a top-down attribution process that grounds a selective set of real instructions to situated users, and 2) a bottom-up synthesis process that leverages web documents to first generate a situation, then a meaningful instruction. This framework allows us to harvest diverse and complex instructions at scale, utilizing the vast range of web documents. Specifically, we construct a dataset of 1 million instructions, called {\SynthQuestionSmallCaps}, and demonstrate that models trained on it achieve leading performance on several common benchmarks, with improvements that continually scale with more web corpora. Data, models and codes will be available at \url{https://github.com/Ignoramus0817/SynthQuestions}.
\end{abstract}

\section{Introduction}
\begin{wrapfigure}{r}{0.45\textwidth}
    \vskip -15pt
    \centering
    \includegraphics[width=0.43\textwidth]{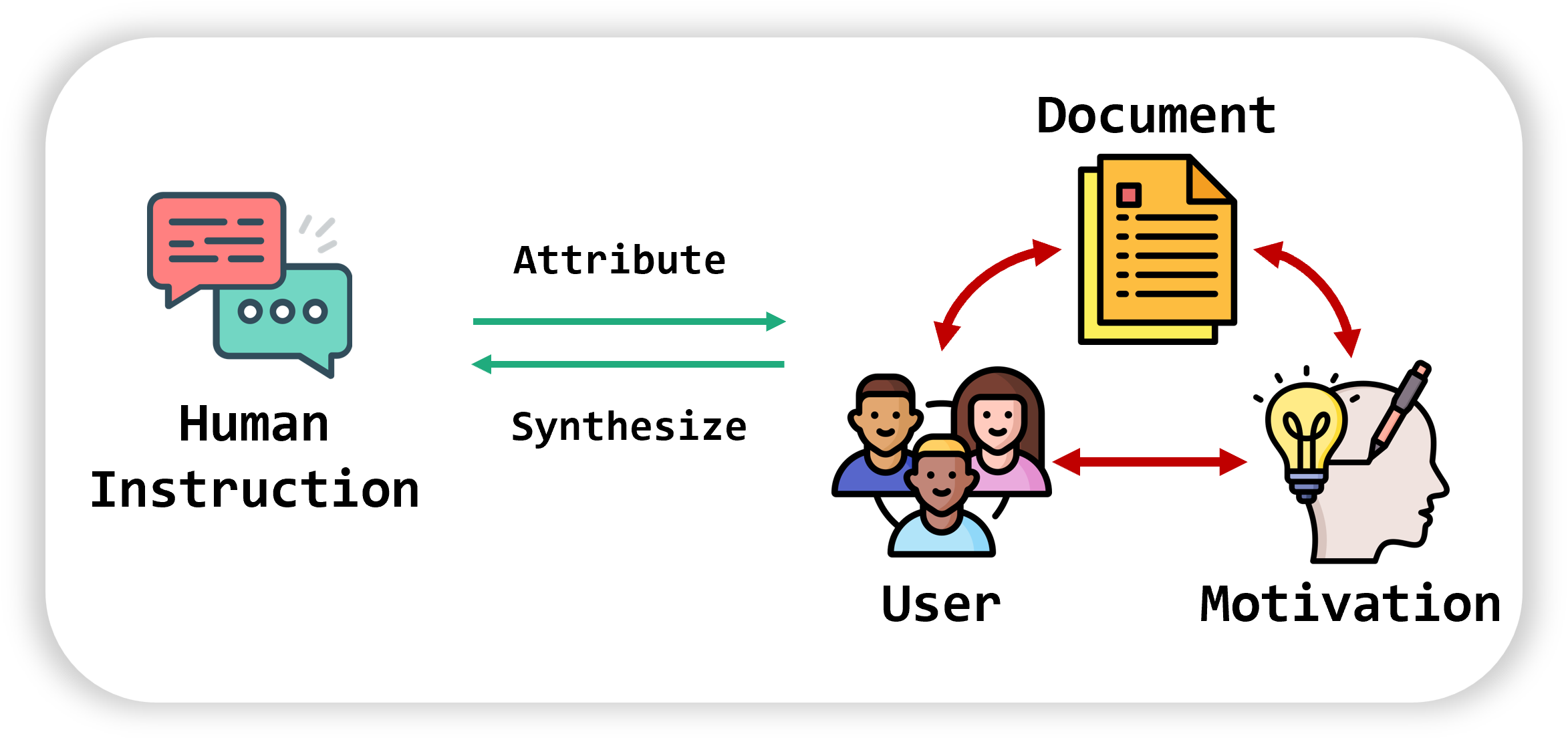}
    \caption{Human instructions can be attributed to documents, users and motivations. Conversely, instructions can also be synthesized from them.}
    \label{fig:attr_schema}
\end{wrapfigure}
Alignment training~\citep{wei2022finetunedlanguagemodelszeroshot} has become an essential technique for instruction-following large language models~\citep{ouyang2022traininglanguagemodelsfollow, openai2024gpt4technicalreport, yang2024qwen2technicalreport, dubey2024llama3herdmodels, glm2024chatglmfamilylargelanguage}, which aims to align language models behaviors with human when given certain instructions through training on instruction-response pairs.
Researchers have been studying how to achieve such alignment effectively, on which educational psychology provides us with wisdom. Vygotsky proposed in his Zone of Proximal Development theory~\citep{vygotsky1978mind} that tasks that are just beyond the learner's capabilities promote maximum cognitive growth, which is suitable for alignment as well.
Showing testimonies to such arguments, plenty of studies have shown that to obtain strong instruction-following and reasoning capability, instruction data that are sufficiently \textbf{diversified}, \textbf{complicated} and \textbf{scaled} are required~\citep{kaplan2020scalinglawsneurallanguage, hoffmann2022trainingcomputeoptimallargelanguage, lu2023instaginstructiontagginganalyzing, köpf2023openassistantconversationsdemocratizing, vicuna2023, zhao2024wildchat1mchatgptinteraction}. 
\footnotetext[4]{Work done during the internship in Metastone Technology.}
\footnotetext[2]{Corresponding author.}

\begin{figure*}[t]
  \centering
  \includegraphics[width=0.85\textwidth]{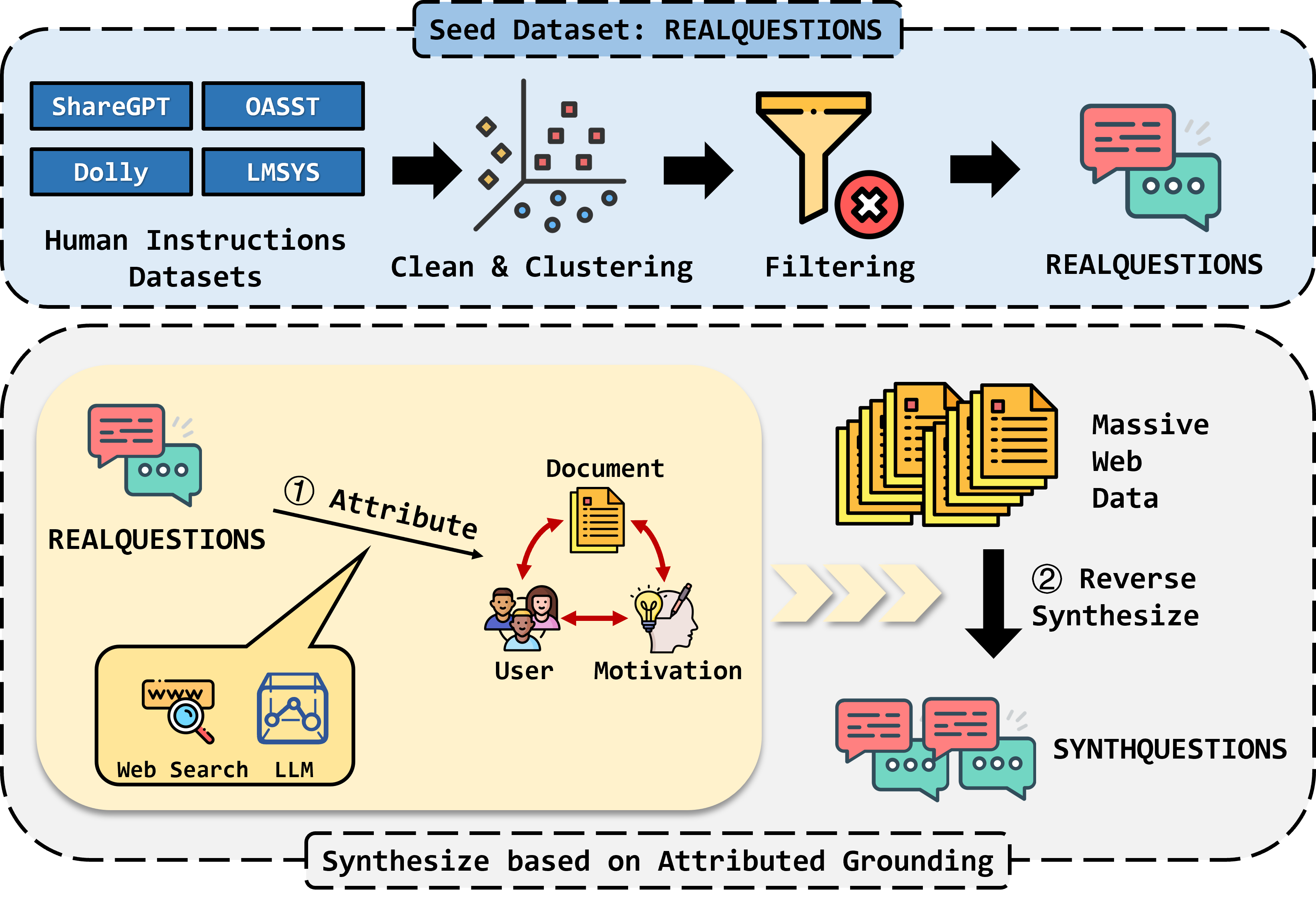}
  \caption{Overview of our synthesizing framework.}
  \label{fig:overview}
\end{figure*}

However, collecting such instructions is an intractable mission, relying on massive use cases and brain labor from human users. As a result, increasing number of works seek to synthesize instructions with language models. Typical existing methods involve augmenting seed tasks~\citep{alpaca, xu2023wizardlmempoweringlargelanguage}, generating instructions according to real-world concepts~\citep{ding2023enhancing}, or training model mimicking human to ask questions~\citep{kong2024platolmteachingllmsmultiround}. While these approaches succeed in the automatic generation of scaled instructional data, they are constrained by the design of their synthesizing methodologies and inevitably fall into certain local distributions (e.g., knowledge and concepts from Wikipedia, limited seed instructional patterns and so on). Consequently, they fall short in generating lifelike and complex instructions that accurately reflect the diversity and intricacy of real-world tasks and queries.

% Generation of instruction data raises us a question: how these instructions come up? As instructions are originally proposed by human, they are invariably originated from certain real-world situations, the breadth of which naturally brings the diversity and complexity to human instructions. As a result, to generate high-quality instructions, it is critical to assure that they are grounded to the world.

% There have been efforts to generate instructions grounded to real-world information. WizardLM~\citep{xu2023wizardlmempoweringlargelanguage} and PlatoLM~\citep{kong2024platolmteachingllmsmultiround} directly use human instructions as grounding contexts by simply modifying them or train a question-raising model on them. Ultrachat~\citep{ding2023enhancing} generates new instructions based on real-world concepts like Wikipedia entities. MAmmoTH2~\citep{yue2024mammoth2scalinginstructionsweb} crawls quiz and exam websites and prompting models to extract Q-A pairs from them. While these methods do leverage real-world information, their grounding processes fail to closely simulate how human instructions are naturally generated, where  users have certain need or faces certain situations. 

Various linguistic and social studies have pointed out that language understanding is based on world knowledge, which is situated, being in part a product of the activity, context, and culture in which it is developed and used~\citep{situated1989, perspective2004, bisk2020experiencegroundslanguage}. Given this idea, we believe it is critical to assure the generated instructions are grounded to real world. In this work, we propose a synthesizing framework based on the core idea of \textbf{attributed grounding}, which consists of two main parts: top-down \textbf{attributing} and bottom-up \textbf{synthesizing}. From our perspective, a human instruction can be attributed to three key factors (as shown in Figure~\ref{fig:attr_schema}): \textbf{(1) Document:} the background knowledge involved in the instruction. \textbf{(2) User:} who proposes this instruction. \textbf{(3) Motivation:} why the users need an LLM to do the task for them. Through these factors, an instruction is grounded to the real world. Conversely, we can build situations including users and motivations from documents and synthesize instructions accordingly. As massive and diverse web documents are accessible without much effort, it is possible to generate pre-training-level instruction data with high complexity and diversity.

\begin{figure*}[t]
    \centering
    \includegraphics[width=0.90\textwidth]{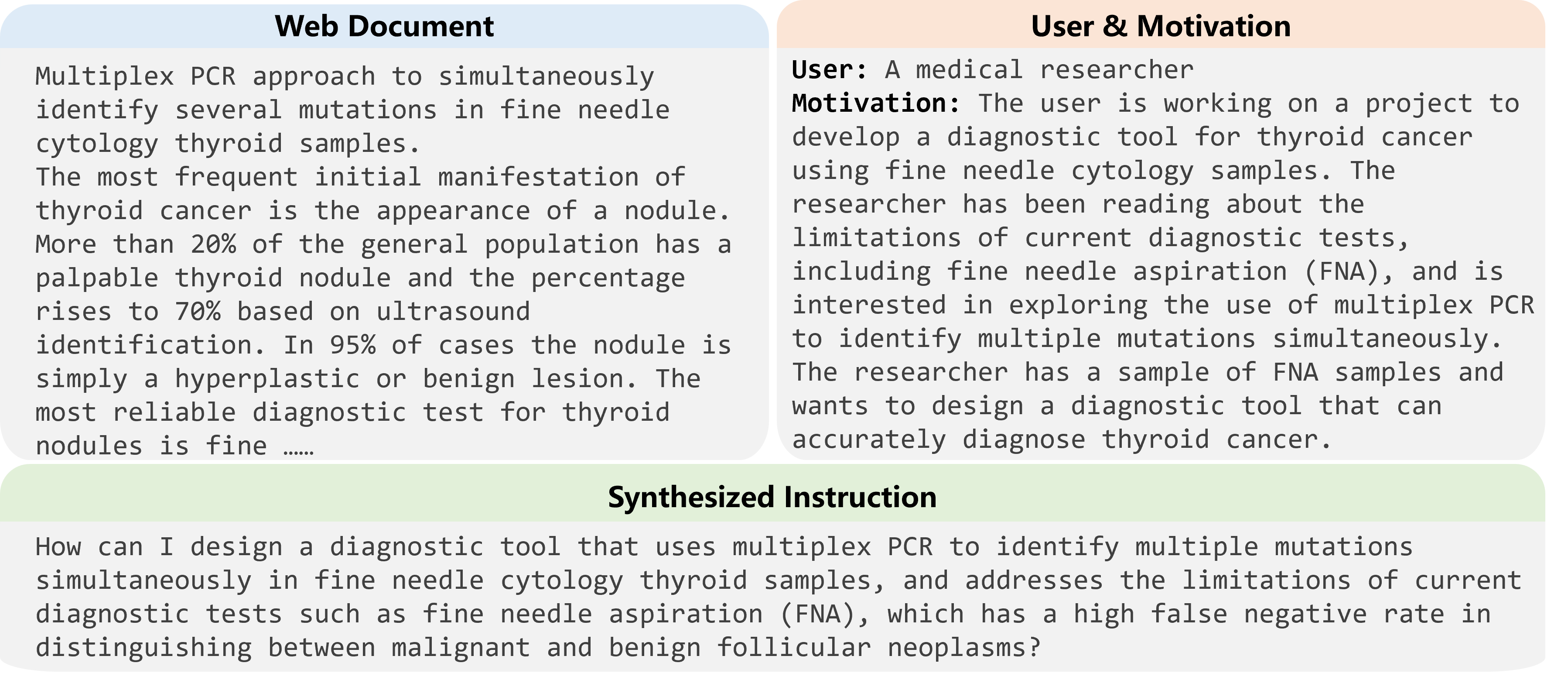}
    \caption{A random sample of web document, 
 the corresponding grounding situation and generated instruction.}
    \label{fig:case}
\end{figure*}

We conduct an in-context-learning driven implementation for our framework, as can be seen in Figure~\ref{fig:overview}. We first build a seed dataset for the attributing process. We collect commonly-used human-labeled SFT datasets, clean and deduplicate the instructions, and keep the ones with the highest quality. We call the resulting seed dataset {\RealQuestionSmallCaps}. In the attributing step, we recall web documents for each instruction in {\RealQuestionSmallCaps}, based on which we build a lifelike situation with users and motivations leveraging an advanced LLM. In the synthesizing step, we start with existing web documents and prompt an LLM to generate grounding situations along with new instructions. Above process is done in an in-context-learning style, driven by demonstrations from {\RealQuestionSmallCaps}. Figure~\ref{fig:case} showcases an example of web document, the corresponding user/motivation and the synthesized instruction, which is well grounded and complicated (see Appendix~\ref{app:cases} for more cases).

With above process we harvest a 1M-size dataset, \textbf{\SynthQuestionSmallCaps}, which exhibits higher diversity than other synthesized datasets. Model fine-tuned on {\SynthQuestionSmallCaps} achieves leading results on various benchmarks and is comparable with models trained with 10 times more data and preference training, which demonstrate the effectiveness of our method.

Our contributions can be concluded as follows:
\begin{itemize}
    \item We construct {\RealQuestionSmallCaps}, an instruction dataset of cleaned and filtered human instructions.
    \item We propose and implement a new data synthesizing framework, which can generate pre-training-scale data with high quality. Based on the framework we construct a 1M scale dataset, \SynthQuestionSmallCaps.
    \item We demonstrate the effectiveness of our methods through intense experiments on series of open-ended and closed-form benchmarks.
\end{itemize}

\section{Related Works}
\subsection{Alignment Training}
Since first appearance in ~\citep{weller-etal-2020-learning}, researchers have found that models can show surprising generalization ability on understanding human intents when fine-tuned with instruction-response pairs~\citep{wei2022finetunedlanguagemodelszeroshot, chung2022scalinginstructionfinetunedlanguagemodels, mishra2022crosstask, ouyang2022traininglanguagemodelsfollow, xu2023expertprompting}. Nowadays supervised fine-tuning (SFT), also referred to as instruction tuning, has become essential for aligning large language models with human behaviors, which is invariably applied to almost every instruction following LLMs~\citep{vicuna2023, openai2024gpt4technicalreport, dubey2024llama3herdmodels, yang2024qwen2technicalreport, glm2024chatglmfamilylargelanguage}. The broad application of SFT raises a high demand on instruction-response data. A number of works have put effort on collecting SFT data from human. ShareGPT~\cite{vicuna2023} and OpenAssistant~\citep{köpf2023openassistantconversationsdemocratizing} collect user conversations with proprietary LLMs like GPT-4. Chatbot Arena~\citep{chiang2024chatbot} is a benchmarking platform for users to chat with different LLMs and rate their responses which collects human instructions in the same time, resulting in several conversation datasets~\citep{zheng2023judging, zheng2024lmsyschat1mlargescalerealworldllm}. However, collecting conversations from users or crowd-sourcing to annotate instruction data is rather expensive, which leads to the limited scale of most human-labeled SFT datasets.

\subsection{Instruction Data Synthesizing}
As the generating ability of LLMs become stronger, more recent works seek to synthesize SFT data automatically to break through the scale limit of human annotating. Self-Instruct~\citep{wang2023selfinstructaligninglanguagemodels} firstly introduces the idea of generating instructions with LLMs themselves. Following this idea,  Alpaca~\citep{alpaca} generates 52K instructions from 175 seed tasks with OpenAI's \texttt{text-davinci-003}. WizardLM~\citep{xu2023wizardlmempoweringlargelanguage} prompts \texttt{gpt-3.5-turbo} to evolve a seed dataset to generate more complicated instructions. 
HumpBack~\citep{li2023self} constructs a translation model that back-translates web documents into instructions.
PlatoLM~\citep{kong2024platolmteachingllmsmultiround} directly train a model on existing SFT datasets to simulate users and raise questions. To improve the diversity and complexity of the instructions, later works begin to inject more real-world information to the generating process. UltraChat~\citep{ding2023enhancing} leverage Wikipedia entities to improve field coverage. MAmmoTH2~\citep{yue2024mammoth2scalinginstructionsweb} directly extract Q-A pairs from web documents and refine them to construct a 10M SFT dataset. 

\section{\RealQuestionSmallCaps}
\label{sec:rq}
{\RealQuestionSmallCaps} is a high-quality human instruction dataset that we construct as the seed dataset, which will later be attributed to grounding situations and drive the synthesizing process as demonstrations. {\RealQuestionSmallCaps} is built with the following steps:    

\paragraph{Data Collection.}We collect conversation data from 7 commonly used human-labeled instruction datasets, namely Chatbot Arena Conversations~\citep{chiang2024chatbot}, Databricks-dolly-15k~\citep{DatabricksBlog2023DollyV2}, LMSYS-Chat-1M~\citep{zheng2024lmsyschat1mlargescalerealworldllm}, OpenAssistant~\citep{köpf2023openassistantconversationsdemocratizing}, ShareGPT~\citep{vicuna2023}, UltraChat~\citep{ding2023enhancing} and WildChat~\citep{zhao2024wildchat1mchatgptinteraction}. We collect a total of 1.92M raw conversation data, details of which are shown in Figure~\ref{tab:data_stat}.

\paragraph{Data Cleaning and Deduplication.}There is massive noise in the raw conversation data and a variety of measures are applied to reduce it:
\begin{itemize}
    \item Conversations that are not complete or do not appear in English are discarded.
    \item Remove conversations where the user instructions are similar to that of our evaluation benchmarks, Alpaca Eval 2.0 and Arena Hard.
    \item As the tasks of user instructions in the conversations exhibit high degree of duplication, we deduplicate them according to their semantics.
\end{itemize}  

\begin{wrapfigure}{r}{0.5\textwidth}
  \includegraphics[width=0.45\textwidth]{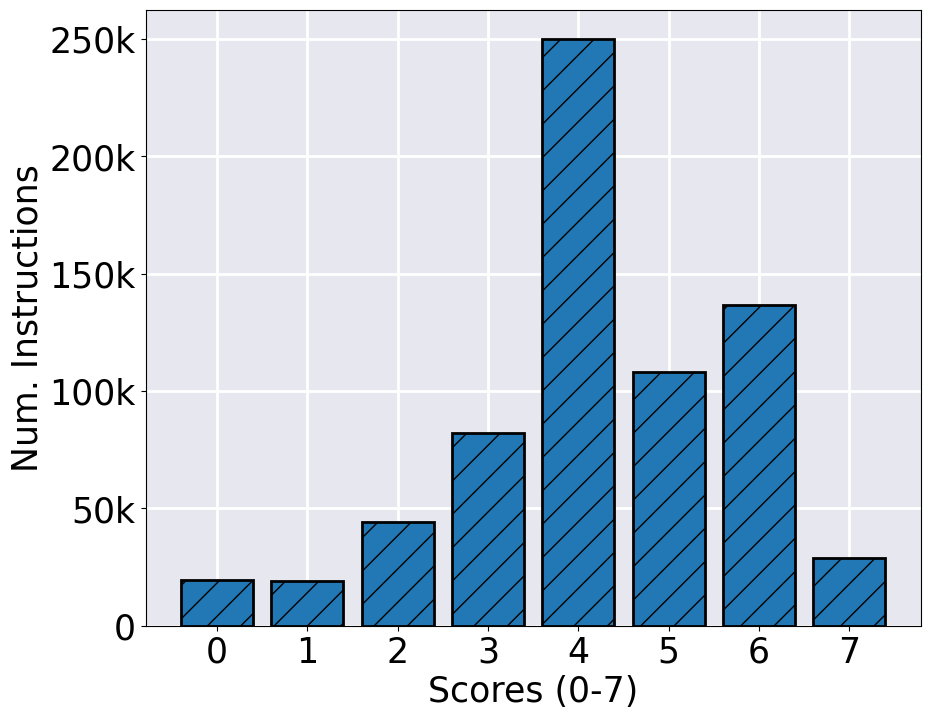}
  \caption{Scores distribution of human instructions, based on Arena Hard Pipeline. The instructions scored 7 go into the {\RealQuestionSmallCaps} dataset.}
  \label{fig:scores}
  \vskip -40pt
\end{wrapfigure}

For the detail of deduplication, we apply a community detection algorithm \footnote{\url{https://sbert.net/docs/package_reference/util.html##sentence_transformers.util.community_detection}} on the instructions, where data points whose embedding similarity exceeds a specified threshold (0.85 in our implementation) are considered to constitute a cluster. For each cluster standing for a certain task or topic, we keep only one instruction to maximize the diversity of the resulted dataset. 
For outlier instructions that do not belong to any clusters, we keep them all as they are not similar to any other ones and can be regarded as unique samples. After the cleaning and deduplication, we get a total of 690K conversations. We discard all the responses from the conversations and only keep the user instructions of the first rounds for later use.

\label{sec:filter}
\paragraph{Data Filtering.}As mentioned previously, challenging tasks are particularly beneficial for model performance. To filter the most challenging instructions, we modify criteria from Arena Hard~\citep{arenahard2024} to score them, where instructions are assessed on 7 dimensions (details in Appendix~\ref{app:arena}). 

We use \texttt{LLaMA-3-70B-Instruct} as the judging model. For each dimension that an instruction satisfies, 1 score will be added. Detailed statistics of scored instructions are shown in Figure~\ref{fig:scores}, where we refer the 29K instructions with full score as our seed dataset \textbf{{\RealQuestionSmallCaps}}.
We fine-tune a \texttt{LLaMA-3-8B} on {\RealQuestionSmallCaps} and experimental results show its superior quality compared to existing instruction datasets (see Table~\ref{tab:rq_res}).

\section{Attributed Grounding}
\subsection{Attributing}
\label{sec:attr}
In the top-down attributing step, instructions from {\RealQuestionSmallCaps} will be attributed to documents, users and motivations. 

\paragraph{Documents.}Attributing starts by collecting documents, i.e. relevant real-world information, for instructions in {\RealQuestionSmallCaps}, which is done with web search in our implementation. We utilize \texttt{LLaMA-3-70B-Instruct} to extract key concepts of each instruction, and recall web pages from Google using the key concepts as queries. We keep the top-1 result as the document for each instruction.
\paragraph{Users and Motivations.}With documents that provide background knowledge about real-world, we can further simulate the situation where the instruction appears. We provide \texttt{LLaMA-3-70B-Instruct} with the document along with the instruction and prompt it to conceive a situation where a user interacts with the document and brings up the instruction out of certain motivation. To improve the grounding process, we conduct the prompting with manually crafted demonstrations, which are shown in Appendix~\ref{app:prompt}. Following above process, we get attributed {\SynthQuestionSmallCaps} which we refer to as:
\begin{equation}
    RQ^\alpha=\{(i, d, u, m)\}
\end{equation}
where i represents instructions from {\SynthQuestionSmallCaps}, $d$, $u$ and $m$ refers to the attributed factors documents, users and motivations respectively.

\subsection{Synthesizing}
In the bottom-up synthesizing process, we reverse the attributing process, constructing situations with users and motivations from existing web documents, based on which new instructions are synthesized. Attributed samples from $RQ^\alpha$ are used here to regulate model behaviors.
\paragraph{Documents.}We use  FineWeb~\citep{penedo2024finewebdatasetsdecantingweb} as the main source for our documents. To further amplify the dataset's benefits for complex reasoning capabilities, we additionally mix in documents that involve difficult reasoning tasks like mathematics and coding from PILE~\citep{gao2020pile800gbdatasetdiverse} and MathPILE~\citep{wang2023generativeaimathi}.

\paragraph{Users and Motivations.}For each document, we prompt \texttt{LLaMA-3-8B-Instruct} to build a grounded situation with users and motivations. To generate more reasonable and grounded situations, samples from $RQ^\alpha$ are used as demonstrations in this process:
\begin{equation}
    (u^\prime, {m}^\prime) = LLaMA(P_g, {d}^\prime, demo)
\end{equation}
where the outputs $u^\prime$ are ${m}^\prime$ are generated user and motivation, the inputs $P$, ${d}^\prime$, and $demo$ are prompt for grounding, documents in the above corpus, and demonstrations from $RQ^\alpha$ respectively.

\paragraph{New Instructions.}Finally we ask the model to play the role of the user and utter the most possible instructions when placed in the above grounding situation:
\begin{equation}
    i^\prime = LLaMA(P_i, {d}^\prime, u^\prime, {m}^\prime, demos)
\end{equation}
where $i^\prime$ is the generated instruction, $P_i$ is the prompt for instruction generation.

After generating new instructions, we first score them with methods in Section~\ref{sec:filter} and discard all the instructions whose scores are below 3 (distribution of the full dataset is shown in Appendix~\ref{app:visual_2}). We choose the threshold 3 as we notice that Specificity, Problem-Solving and Technical Accuracy are three more fundamental requirements for a valid instruction, and instructions failing these three requirements tend to be unclear or ambiguous. As this is a heuristic setting, we set the threshold 3 instead of directly filtering out instructions that do not meet these requirements to be more tolerant. To assure the diversity of the dataset, we conduct topic modeling with BERTopic~\citep{grootendorst2022bertopic} following~\citep{arenahard2024} and construct a 1M-size final dataset by including instructions with the highest scores in each topic. We refer to the final dataset as \textbf{{\SynthQuestionSmallCaps}}.

\begin{figure}[t]
    \includegraphics[width=0.418\columnwidth]{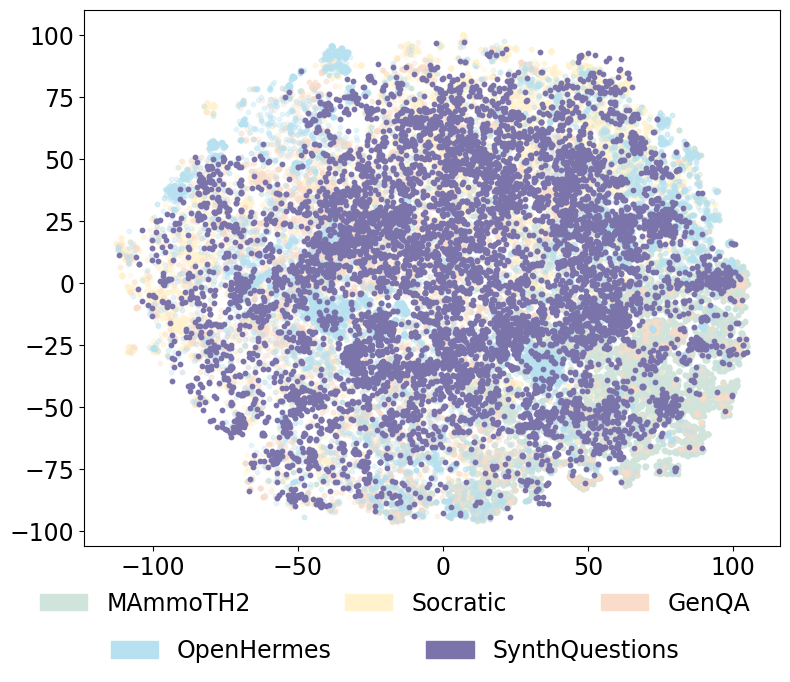}
    \includegraphics[width=0.542\columnwidth]{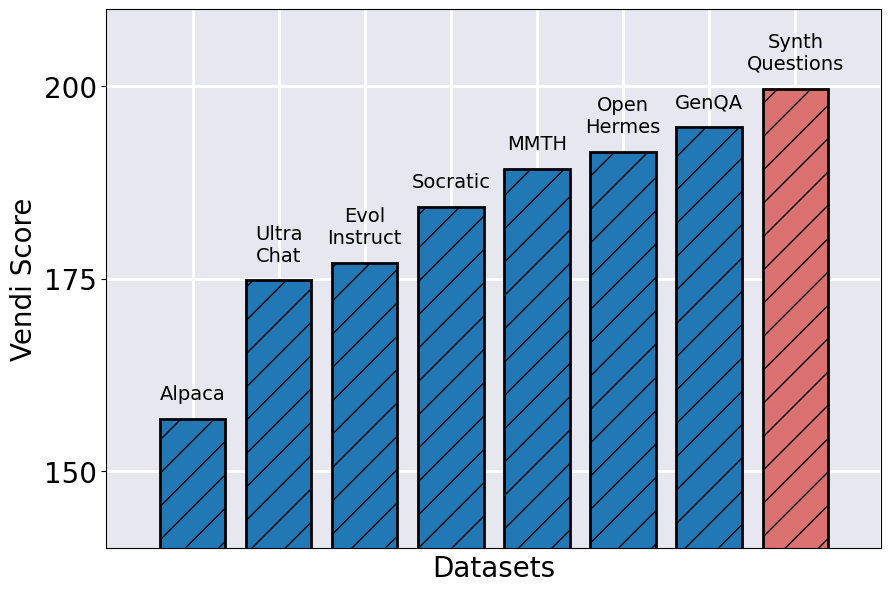}
    \label{fig:tsne_vendi}
    \caption{\textbf{Left}: t-SNE plot of {\SynthQuestionSmallCaps} along with MAmmoTH2, SocraticChat, GenQA and OpenHermes. {\SynthQuestionSmallCaps} covers more area than other datasets. \textbf{Right}: Comparison of diversity (vendi score) of synthesized datasets.}
\end{figure}

\section{Data Analysis}
In this section we demonstrate the diversity and complexity of {\SynthQuestionSmallCaps} through qualitative and quantitative evaluations.
\subsection{Basic Statistics}
Table~\ref{tab:data_stat} in Appendix~\ref{app:data_stat} shows the basic statistics of {\SynthQuestionSmallCaps} along with other instruction datasets(tokenization is done with Tiktoken\footnote{\url{https://github.com/openai/tiktoken}}). As can be seen in the table, our method generates data with the most average turn lengths, indicating the complexity of {\SynthQuestionSmallCaps}. Besides, {\SynthQuestionSmallCaps} ranks top in the lexical diversity calculated with MTLD algorithm~\citep{mccarthy2010mtld} among all synthesized datasets.

\subsection{Diversity}
Apart from lexical diversity in Table~\ref{tab:data_stat}, we assess and visualize the semantic diversity of instructions in {\SynthQuestionSmallCaps} along with other synthetic datasets. We sample 10,000 instructions from each dataset above and calculate sentence embeddings with \texttt{all-mpnet-base-v2} model\footnote{\url{https://huggingface.co/sentence-transformers/all-mpnet-base-v2}}. Then t-SNE is applied to project semantic embeddings into a 2D space. Figure~\ref{fig:tsne_vendi} display the t-SNE plots of the five most recent datasets(more datasets are visualized in Appendix~\ref{app:visual_2}), where data points from {\SynthQuestionSmallCaps} occupy the most extensive area. This implies that dataset synthesized with our method covers more diverse topics or subjects.

\begin{wrapfigure}{r}{0.45\textwidth}
    \centering
    \includegraphics[width=0.43\textwidth]{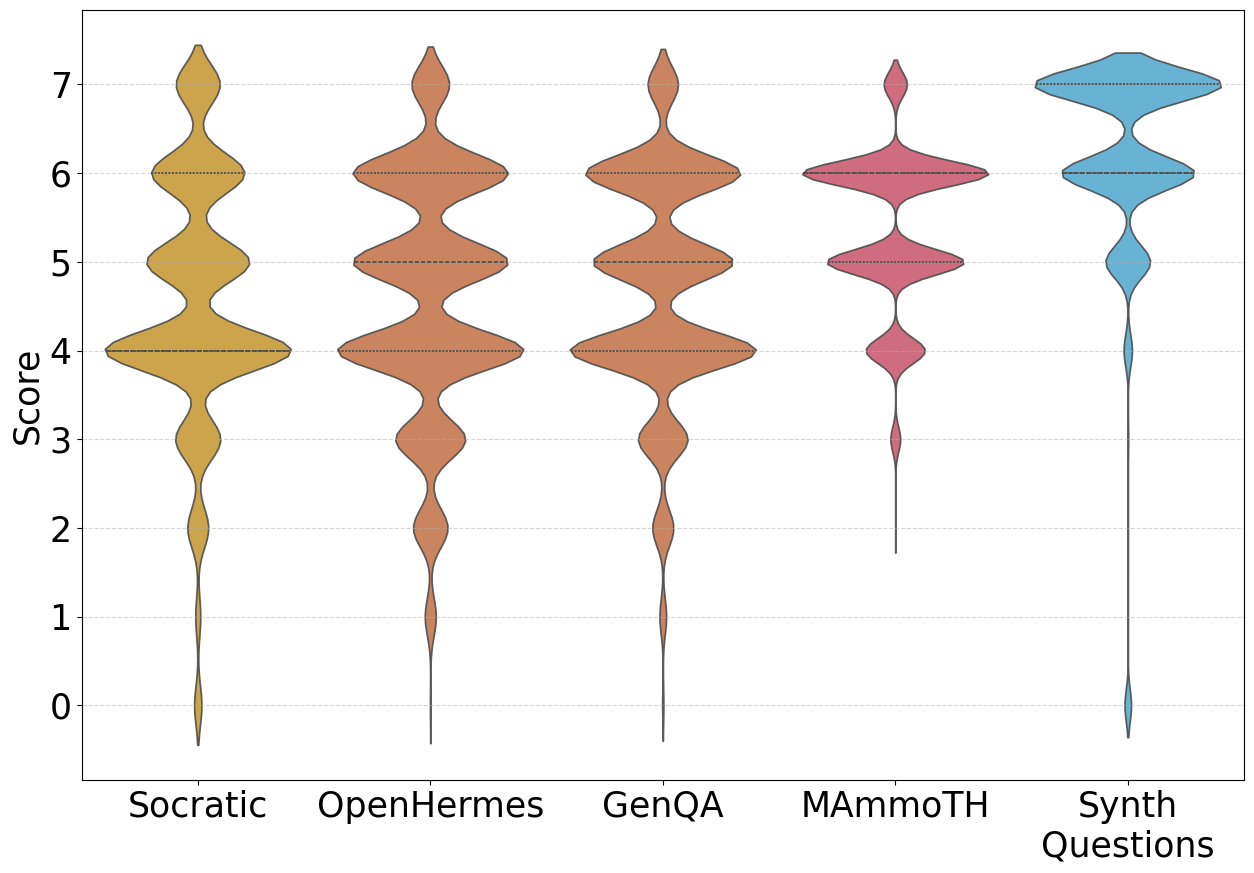}
    \caption{Comparison of complexity (Arena Hard Score) of all synthetic data.}
    \label{fig:score_violin}
    \vskip -40pt
\end{wrapfigure}

Additionally, we provide Vendi Score as a supplemental quantitative metric of data diversity. Vendi Score~\citep{friedman2023vendiscorediversityevaluation} is a diversity evaluation metric designed for machine learning dataset, which increases linearly with the number of unique modes in the dataset. We compute Vendi Score on the previously sampled 10,000 data points for each dataset. Results are presented in Figure~\ref{fig:tsne_vendi}, where Vendi Score of {\SynthQuestionSmallCaps} ranks top among all synthesized datasets.

\subsection{Complexity}
We randomly sample 10,000 unfiltered instructions from all the synthetic datasets in Table~\ref{tab:data_stat} and score them with criteria from Arena Hard. Results in Figure~\ref{fig:score_violin} display the violin plots of score distribution of the 5 most recent dataset(full results in Appendix~\ref{app:visual_2}), where the score distribution of {\SynthQuestionSmallCaps} is notably condensed to the higher end of spectrum compared to other datasets. Such observation signifies the efficiency in our framework to produce complicated instructions.

\subsection{Safety}
We analyze \SynthQuestionSmallCaps with LLaMA-Guard-3-8B, detecting potential harmful content in the dataset (results shown in Appendix~\ref{app:safety}). Among the 1M dataset, we detected 4.32\% data points with potential safety risks. Detailed results are shown below, where most of problematic data points (3.60\%) lie in the "Specialized Advice" category, which we think are the ones requiring users to carefully discern the responses, instead of being directly harmful. For all the other categories, the potentially risky data points are less than 0.2\%. When open-sourcing we will remove contents with potential harm.

\begin{table*}[t]
  \centering
  \resizebox{\textwidth}{!}{
  \begin{tabular}{clccccc}
    \hline
    \multirow{2}{*}{\textbf{Model Groups}} &\multirow{2}{*}{\textbf{Models}} & \textbf{Data} & \textbf{Arena Hard} & \multicolumn{3}{c}{\textbf{Alpaca Eval 2.0}} \\
    & & \textbf{Scale} & WR(\%) & LC(\%) & WR(\%) & SD \\
    \hline
    \multirow{3}{*}{\shortstack{Larger or \\ Proprietary Models}}  & HumpBack-LLaMA2-70B & - & -  & 16.25 & 10.12 & 0.94\\
    & GPT-3.5-Turbo-0301 & - & 18.1  & 18.09 & 9.62 & 0.89      \\
    & GPT-3.5-Turbo-1106 & - & 18.9  & 19.30 & 9.18 & 0.89      \\
    \hline
    \multirow{3}{*}{\shortstack{7B-8B Models \\ w/ Proprietay Data}} & Mistral-7B-Instruct-v0.3 & - &  16.7$^{\dag}$  & 20.61 & 16.69 & 1.11 \\
    % & Qwen-1.5-7B-Chat & 14.75 & 11.77 & 0.95 & -    \\
    % & OpenHermes(Mistral)  & 16.25 & 10.34 & 0.94 & -    \\
    & Qwen-2-7B-Insturct$^{\dag}$ & - & 23.5 & 21.86 & 19.62 & 1.15    \\
    & LLaMA-3-8B-Instruct & >10M  & 20.6 & 22.92 & 22.57 & 1.26  \\
    \hline
    \multirow{3}{*}{\shortstack{LLaMA3-8B \\w/ Open-source Data}} & OpenHermes2.5 & 1M & 4.4 & 9.94 & 6.27 & 0.73 \\
    & GenQA & 10M & 3.0 & 9.05 & 7.11 & 0.82 \\
    & MAmmoTH2$^*$ & 10M & \textbf{16.6} & \underline{18.5} & - & -\\
    \hline
    Ours & \textbf{{\SynthQuestionSmallCaps}} & \textbf{1M} & \underline{15.4} & \textbf{18.87} & \textbf{19.15} & 1.15 \\
    \hline
  \end{tabular}
  }
  \caption{Performance of models on Alpaca Eval 2.0 and Arena Hard benchmarks. Among models with open-source data, the best performance is bolded and the second best performance is underlined. Results marked with \dag\; are evaluated by us. *Apart from synthetic data, MAmmoTH2 is further fine-tuned with external math and code datasets, which may explain its high performance on Arena Hard.}
  \label{tab:chat_performance}
\end{table*}

\begin{table*}[t]
  \centering
  \resizebox{\textwidth}{!}{
  \begin{tabular}{clcccccc}
    \hline
    \textbf{Model Groups} & \textbf{Models} & \textbf{IFEVAL} & \textbf{MMLU} & \textbf{ARC-C} & \textbf{GPQA} & \textbf{GSM8K} & \textbf{MATH} \\
    \hline
    \multirow{3}{*}{\shortstack{7B-8B Models \\ w/ Proprietay Data}} & Mistral-7B-Instruct-v0.3 & 54.65 & 61.84 & 63.57 & 27.8$^{\dag}$ & 43.37 & 12.94$^{\dag}$ \\
    & Qwen-2-7B-Insturct & 56.79 & 70.5 & 59.73 & 25.3 & 82.3 & 49.6 \\
    & LLaMA-3-8B-Instruct  & 74.08 & 68.5 & 82.4 & 34.6 & 80.6 & 29.1 \\
    \hline
    \multirow{3}{*}{\shortstack{LLaMA3-8B \\w/ Open-source Data}} & OpenHermes2.5 & - & \underline{65.7} & 61.86 & - & 67.02 & - \\
    & GenQA & - & 63.45 & 58.53 & - & 43.13 & - \\
    & MAmmoTH2 & 43.94$^{\dag}$ & 64.2 & \textbf{82.2} & \textbf{35.2} & \underline{70.4} & \textbf{35.8} \\
    \hline
    Ours & \textbf{{\SynthQuestionSmallCaps}} & \textbf{57.05} & \textbf{65.79} & \underline{63.92} & \underline{30.3} & \textbf{70.53} & \underline{22.71} \\
    \hline
  \end{tabular}
  }
  \caption{Performance of models on different close-ended knowledge and reasoning benchmarks. Notations and marks are the same with the above figure. Some unreported results are not reproduced due to high expense.}
  \label{tab:close_benchmarks}
\end{table*}

\section{Experiments}
In this section, we first verify that data synthesized with our method can improve model's instruction following and reasoning performance effectively. Then we show the scaling curve of our synthesized data, demonstrating how model performance changes as data scale increases. Finally through ablation experiments we demonstrate the potential of our data to be further elicited by preference optimization and the necessity of each module in our method. Instruction tuning and preference optimization are repectively conducted with Megatron-LM\footnote{\url{https://github.com/NVIDIA/Megatron-LM}} and Huggingface TRL\footnote{\url{https://huggingface.co/docs/trl/index}}. Full training details are shown in Appendix~\ref{app:train}.

\subsection{Main Results}
Due to page limit, we omit the evaluation results of datasets that are relatively old and less competitive (e.g. UltraChat, ShareGPT and so on). Complete results for them can be found in Appendix~\ref{app:full_eval}.

\paragraph{Alignment Benchmarks.} To verify the effectiveness of our method and the quality of synthesized data, we train \texttt{LLaMA-3-8B} on {\SynthQuestionSmallCaps}. For evaluation, we select two prevailing alignment benchmarks, Alpaca Eval 2.0~\citep{dubois2024lengthcontrolledalpacaevalsimpleway} and Arena Hard~\citep{arenahard2024}, which leverage \texttt{gpt-4-1106-preview} as the judge and are highly consistent with human annotation.

Results are shown in Table~\ref{tab:chat_performance}. Among all models trained on open-source datasets, model trained on {\SynthQuestionSmallCaps} shows the best performance on Alpaca Eval 2.0 and only falls behind \texttt{MAmmoTH2} on Arena Hard. It is especially worth mentioning that on Alpaca Eval 2.0, model trained with {\SynthQuestionSmallCaps} outperforms ones trained with \texttt{MAmmoTH2} and \texttt{HumpBack}, which are two synthesizing methods also utilizing web documents. Note that \texttt{MAmmoTH2} is trained with 10M data and further fine-tuned with open-source math and code datasets, which may explain its high win rate on Arena Hard. Models trained with {\SynthQuestionSmallCaps} is also comparable to latest models trained with proprietary data and reinforcement learning, showing a better win rate on Alpaca Eval 2.0 than \texttt{Mistral-7B-Instruct-v0.3}. Above experimental results prove the effectiveness of {\SynthQuestionSmallCaps} on improving models capabilities.

\paragraph{Closed-form Benchmarks.} To demonstrate the robustness of our dataset, we evaluate the models on several closed-form benchmarks. We present the results in Table~\ref{tab:close_benchmarks}, where model trained on {\SynthQuestionSmallCaps} ranks either first or second on all benchmarks among models trained with open-source data.

\paragraph{Preference Optimization.} We further investigate the potential of our synthesized dataset by applying DPO. We randomly sample 100K instructions from different datasets, generate 5 responses(T=0.8) with \texttt{LLaMA-3-70B-Instruct}, and label the preferences with \texttt{ArmoRM-Llama3-8B-v0.1}~\citep{ArmoRM}. We set the response with the highest score as the chosen, and the one with the lowest score as the rejected. We train our the SFT model and the performance of resulted model are presented in~\ref{tab:dpo}. Model trained on {\SynthQuestionSmallCaps} not only outperforms all recent synthesized datasets, but also beats \texttt{LLaMA-3-8B-Instruct}. Our model even surpasses the data generator \texttt{LLaMA-3-70B-Instruct} on Alpaca Eval 2.0 Win Rate. However, the performance on Arena Hard falls far behind, which may indicate that solving more difficult tasks still calls for an increase in model scales.

\begin{table}[t]
\begin{minipage}[b]{0.48\linewidth}
  \centering
  \begin{tabular}{lcc}
    \hline
    \multirow{2}{*}{\textbf{Models}} & \textbf{Alpaca} & \textbf{Arena} \\
    & \textbf{Eval (WR)} & \textbf{Hard} \\
    \hline
    LLaMA-3-8B-Instruct & 22.56 & 20.6 \\
    LLaMA-3-70B-Instruct & 33.18 & 44.5 \\
    \hline
    {\SynthQuestionSmallCaps} & 19.15 & 15.4 \\
    \quad +DPO (MAmmoTH2) & 28.46 & 15.6 \\
    \quad +DPO (GenQA) & 28.52 & 17.4 \\
    \quad +DPO (OpenHermes) & \underline{28.94} & \underline{19.6} \\
    \quad\textbf{+DPO (Ours)} & \textbf{33.81} & \textbf{24.8} \\    
    \hline
  \end{tabular}
  \vskip 10pt
  \caption{Performance of models with DPO on different datasets. Our DPO model even outperforms LLaMA-3-70B-Instruct on Alpaca Eval 2.0 Win Rate.}
  \label{tab:dpo}
\end{minipage}\hfill
\begin{minipage}[b]{0.48\linewidth}
    \centering
    \includegraphics[width=0.8\linewidth]{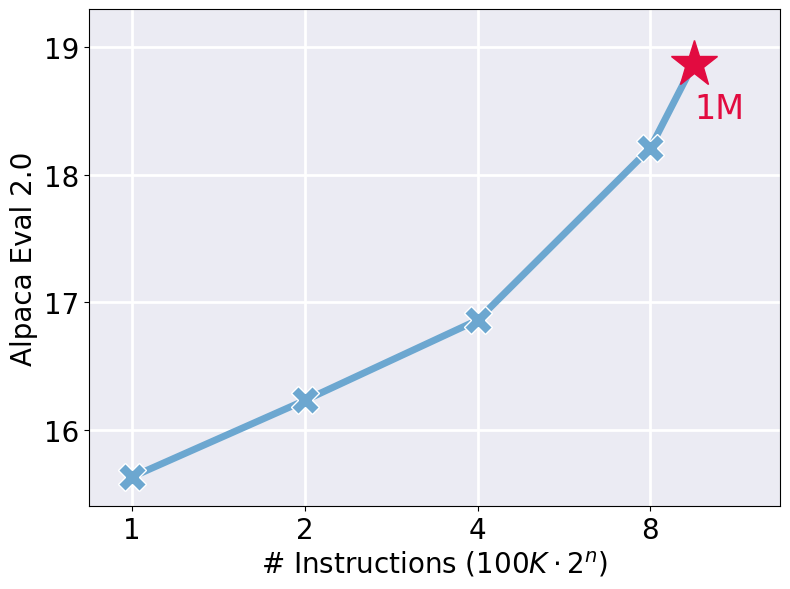}
    \captionof{figure}{Performance of models trained with subsets in different scales.}
    \label{fig:scaling}
\end{minipage}
\end{table}

\subsection{Scaling Effect}
We investigate the impact of instruction scale to model performance. We randomly draw $2^i\cdot100$K subsets from the total 1M data and train \texttt{LLaMA-3-8B} models on them. Performance of the models are evaluated with Alpaca Eval 2.0, results of which are displayed in Figure~\ref{fig:scaling}. As the scale of the train set increases, model performance consistently improves on the benchmark, which corroborates the positive impact of expanding the scale of data on enhancing model performance, while also demonstrating the potential of our approach to continuously improve model capabilities by synthesizing more instructions. We also provides results of Qwen-2.5-7B trained on subsets with different scales in Appendix~\ref{app:qwen_res}, which shows similar trends.

\subsection{Performance on Model of Other Structures and Sizes}
To verify the generalizability of \SynthQuestionSmallCaps, we train another two models, \texttt{Qwen2.5-7B} and \texttt{Qwen2.5-14B} on a 100k-subset randomly drawn from different synthesized datasets including OpenHermes2.5, GenQA, MAmmoTH2 and \SynthQuestionSmallCaps. We evaluate the model performance on Alpaca Eval 2.0 and closed-form knowledge and reasoning benchmarks. As is shown in Table~\ref{tab:other_models}, models trained with {\SynthQuestionSmallCaps} ranks first on most of the benchmarks with a non-trivial advantage. Especially, in MATH dataset, model trained with {\SynthQuestionSmallCaps} outperforms other models by a large margin. We also notice that {\SynthQuestionSmallCaps} does not perform as strong on GPQA, which may indicate that it still requires enhancement in subjects like biology, physics, and chemistry. Nevertheless, present results can already demonstrate the effectiveness and generalizability of our method.

\begin{table*}[t]
  \centering
  \resizebox{\textwidth}{!}{
  \begin{tabular}{clcccccccc}
    \hline
    \textbf{Model} & \textbf{Dataset} & \textbf{AE (LC)} & \textbf{AE (WR)} & \textbf{IFEVAL} & \textbf{MMLU} & \textbf{ARC-C} & \textbf{GPQA} & \textbf{GSM8K} & \textbf{MATH} \\
    \hline
    \multirow{4}{*}{Qwen2.5-7B} & OpenHermes2.5 & \underline{16.09} &	\underline{9.85} &	\underline{48.08} &	68.86 &	82.17 &	\underline{32.32} &	\underline{78.62} &	29.8 \\
    & GenQA & 11.33 &	6.23 &	42.21 &	67.49 &	84.64 &	\textbf{33.33} &	72.63 &	29.74 \\
    & MAmmoTH2  & 11.38 &	6.72 &	41.61 &	\underline{69.68} &	\underline{86.95} &	25.76 &	68.02 &	\underline{35.48} \\
    & \textbf{\SynthQuestionSmallCaps}  & \textbf{17.25} &	\textbf{16.03} &	\textbf{48.68} &	\textbf{70.2} &	\textbf{87.88} &	28.12 &	\textbf{79.91} &	\textbf{41.22} \\
    \hline
    \multirow{4}{*}{Qwen2.5-14B} & OpenHermes2.5 & \textbf{24.89} &	\underline{13.21} &	\underline{51.02} &	74.37 &	89.16 &	32.32 &	\underline{85.14} &	36.82 \\
    & GenQA & 12.06 &	6.31 &	42.03 &	72.98 &	89.93 &	\textbf{38.38} &	77.48 &	37.66 \\
    & MAmmoTH2  & 16.96 &	8.30 &	44.18 &	\textbf{77.2} &	\underline{90.44} &	32.83 &	79.38 &	\underline{40.48} \\
    & \textbf{\SynthQuestionSmallCaps}  & \underline{24.33} &	\textbf{22.22} &	\textbf{58.03} &	\underline{77.16} &	\textbf{90.61} &	\underline{34.34} &	\textbf{87.49} &	\textbf{44.96} \\
    \hline
  \end{tabular}
  }
  \caption{Performance of \texttt{Qwen2.5-7B} and \texttt{Qwen2.5-14B} trained on different 100K datasets. AE denotes Alpaca Eval 2.0}
  \label{tab:other_models}
\end{table*}

\subsection{Ablations}
In this section, we verify the necessity of each module by ablate certain parts of our synthesizing framework. Figure ~\ref{tab:ablation} presents the main results of our ablation study.

\paragraph{Attributed Grounding.} We test the effect of our core idea, i.e. attributed grounding, by directly generating instructions without attributing them to documents, users or motivations. We leverage instructions from {\RealQuestionSmallCaps} as demonstrations and prompt \texttt{LLaMA-3-70B-Instruct} to generate new ones. When selecting demonstrations, we apply two strategies: randomly sampling and semantics-based selecting. For the latter strategy we randomly sample an instruction from {\RealQuestionSmallCaps} and then search for K nearest instructions from the whole unfiltered human instruction set (mentioned in Section~\ref{sec:rq}). Detailed prompts are shown in Appendix~\ref{app:prompt}. We collect 100K instructions for each strategy and train models on them.

As shown in the table, data generated with both strategy behave similarly, bringing little improvement to model performance compared to {\SynthQuestionSmallCaps}-100K. Such degeneration demonstrate the effectiveness of attributing process in our method, underlying the critical role of grounding.

\begin{wrapfigure}{r}{0.45\textwidth}
  \centering
  \begin{tabular}{lcc}
    \hline
    \textbf{Training Set} & \textbf{AE} & \textbf{GSM8K}\\
    \hline
    {\SynthQuestionSmallCaps}-100K & 15.63 & 58.30 \\    
    \hline
    \textbf{- Attributed Grounding} \\
    \quad w/ KNN        & 10.85  & 45.26  \\
    \quad w/o KNN        & 10.64 & 43.77  \\
    \hline
    \textbf{- Math/Code Docs} &  15.50 & 50.34 \\
    \hline
  \end{tabular}
  \captionof{table}{Results of ablation study on grounding and additional documents. AE denotes Alpaca Eval 2.0.}
  \label{tab:ablation}
  \vskip -10pt
\end{wrapfigure}

\paragraph{Math and Code Documents.} Apart from FineWeb, we add documents involving more difficult tasks like math or code when synthesizing new instructions. To study the effect of these documents, we randomly sample 100K instructions purely generated from FineWeb and train a model on it. Results in Table~\ref{tab:ablation} show that though removing additional documents does not cause significant performance degradation on Alpaca Eval 2.0, the accuracy on GSM8K drops severely. Such phenomenon reveals that incorporating instructions covering more challenging tasks or fields are especially beneficial for model performance on reasoning and knowledge tasks. 

\section{Conclusion}
In this paper, we propose a two-step instruction synthesizing framework aimed at generating better grounded instruction data. Our framework first attributes human instructions to documents, users and motivations, and then reversely generate grounded instructions from existing web documents through simulating the natural appearance of human instructions. With our synthesizing framework, we construct {\SynthQuestionSmallCaps}, a 1-million synthesized instruction dataset. We fine-tuned LLaMA-3-8B models on our synthesized data and experiments shown that {\SynthQuestionSmallCaps} can enhance model capabilities effectively, achieving comparable performance with models trained with 10 times more data and preference training. Apart from decent performance, study about the scaling effect of {\SynthQuestionSmallCaps} demonstrates the potential of our method to further improve model capabilities by synthesizing larger scales of data.

\section*{Acknowledgment}
This research is supported by Artificial Intelligence-National Science and Technology Major Project 2023ZD0121200 and  National Natural Science Foundation of China under Grant 62222212.

\section*{Limitation and Potential Risks}
The main limitations of this work fall in two aspects. Firstly, while scaling curve shows the potential to further improve model performance, we do not test data scale larger than 1M. Secondly, while it is not the main topic of this work, a more thorough study about the optimal selection and distribution of web corpora used for synthesizing can be conducted.
For risks, the dataset has not been assessed in terms of hallucination, which may lead language models to output false or unfaithful contents.

\bibliographystyle{plain} 
\bibliography{custom}

\clearpage
\onecolumn
\appendix
\section{Dataset and Model Licenses}
Here are the licenses of datasets and models used in this paper:
\begin{itemize}
    \item \textbf{CC}: Chatbot Arena Conversations
    \item \textbf{CC-BY-SA-3.0}: Databricks-Dolly-15K
    \item \textbf{Apache-2.0}: OpenAssistant, ShareGPT
    \item \textbf{ODC-by}: WildChat
    \item \textbf{Others}: InstructionWild (Non-commercial Use), LLaMA-3 family (customized license), LMSYS-Chat-1M (Unknown), 
\end{itemize}

\section{Performance on {\RealQuestionSmallCaps}}
Below is the performance of LLaMA-3-8B trained on different human instruction datasets, where {\RealQuestionSmallCaps} ranks first.

\begin{table}[h]
  \centering
  \begin{tabular}{lr}
    \hline
    \textbf{Training Set} & \textbf{Alpaca Eval 2.0} \\
    \hline
    OASST                        & 4.51       \\
    Chatbot-Arena Convs.         & 5.17       \\
    UltraChat                    & 6.20       \\
    ShareGPT                     & 9.13       \\
    WildChat                     & 14.62      \\
    \hline
    {\RealQuestionSmallCaps}     & \textbf{16.77}     \\
    \hline
  \end{tabular}
  \caption{Performance of models trained on different datasets evaluated with Alpaca Eval 2.0.}
  \label{tab:rq_res}
\end{table}

\section{Training Details}
\label{app:train}
Table below is the hyper-parameters used in finetuning models. For SFT, we use Megatron-LM and 8*8*Nvidia H100 GPUs. For DPO, we use Huggingface TRL and 8*Nvidia H100 GPU.

\begin{table}[h]
\begin{minipage}{.5\columnwidth}
\centering
\begin{tabular}{lr}
\hline
\textbf{Parameters} & \textbf{Values} \\
\hline
Epoch      & 3 \\
Learning Rate & $2e^{-5}$ \\
Global Batch Size & 128 \\
Gradient Accumulation & 1 \\
Gradient Checkpointing & False \\
Precision  & BF16 \\    
Max Length & 8192 \\
Warmup Ratio & 0.06 \\
Weight Decay & 0 \\
Learning Rate Scheduler & Cosine \\
\hline
\end{tabular}
\caption{Hyper-parameters of SFT.}
\end{minipage}
\begin{minipage}{.5\columnwidth}
\centering
\begin{tabular}{lr}
\hline
\textbf{Parameters} & \textbf{Values} \\
\hline
Epoch      & 1 \\
Learning Rate & $0.7e^{-6}$ \\
Global Batch Size & 128 \\
Gradient Accumulation & 8 \\
Gradient Checkpointing & True \\
Precision  & BF16 \\    
Max Length & 8192 \\
Warmup Ratio & 0.1 \\
Weight Decay & 0 \\
Learning Rate Scheduler & Cosine \\
\hline
\end{tabular}
\caption{Hyper-parameters of DPO.}
\end{minipage}
\quad

\end{table}

\clearpage
\section{Statistics of Datasets}
\label{app:data_stat}
Below is the basic statistics of common instructional datasets, where \SynthQuestionSmallCaps ranks top in tokens per turn and lexical diversity.
\begin{table*}[h]
  \centering
  \begin{tabular}{clcccc}
    \hline
    \textbf{Source} & \textbf{Dataset Name} & \#\textbf{Convs} & \#\textbf{Turns} & \#\textbf{Tokens/T} & \textbf{Lex-Div}  \\
    \hline
    \multirow{7}{*}{Human} & ShareGPT~\citep{vicuna2023} & 62K & 3.42 & 335 & 59.86   \\
    & Chatbot Arena Convs~\citep{chiang2024chatbot} & 33K & 1.19 & 224 & 58.44 \\
    & InstructionWild~\citep{instructionwild} & 110K & 1 & 76 & 80.18\\
     & OpenAssistant~\citep{köpf2023openassistantconversationsdemocratizing} & 48K & 1.73 & 211 & 69.64   \\
     & WildChat~\citep{zhao2024wildchat1mchatgptinteraction} & 652K & 2.52 & 519 & 86.58   \\
     & Databricks-Dolly~\citep{DatabricksBlog2023DollyV2} & 15K & 1 & 174 & 76.62 \\
     & LMSYS-Chat-1M~\citep{zheng2024lmsyschat1mlargescalerealworldllm} & 1M & 2.01 & 248 & 59.94  \\
    \hline
    \multirow{8}{*}{Synthetic} & Alpaca~\citep{alpaca} & 52K & 1 & 74 & 64.08   \\
    & UltraChat~\citep{ding2023enhancing} & 208K & 3.16 & 364 & 73.90 \\
    & Evol Instruct~\citep{xu2023wizardlmempoweringlargelanguage} & 143K & 1 & 475 & 60.19 \\
    & SocraticChat~\citep{kong2024platolmteachingllmsmultiround} & 50K & 5.28 & 345 & 65.67 \\
    % & MagPie~\cite{xu2024magpiealignmentdatasynthesis} & 300K & 1 & 680 & \textbf{84.22} \\
    & OpenHermes~\cite{OpenHermes2.5} & 1M & 1 & 346 & 60.10 \\
    & MAmmoTH2~\cite{yue2024mammoth2scalinginstructionsweb} & 10M & 1 & 331 & 53.00 \\
    & GenQA~\cite{chen2024genqageneratingmillionsinstructions} & 11M & 1.69 & 167 & 58.75\\
    & \textbf{{\SynthQuestionSmallCaps}} & 1M & 1 & \textbf{802} & \underline{77.19} \\
    \hline
  \end{tabular}
  \caption{Statistics of different datasets. \#Tokens/T represents number of tokens per turn. Lexical diversity is calculated with 10,000 samples randomly drew from the full dataset.}
  \label{tab:data_stat}
\end{table*}

\section{Supplemental Visualization of Dataset Diversity and Complexity}
\label{app:visual_2}
\begin{figure}[h]
\begin{minipage}{.3\columnwidth}
    \centering
    \includegraphics[width=\columnwidth]{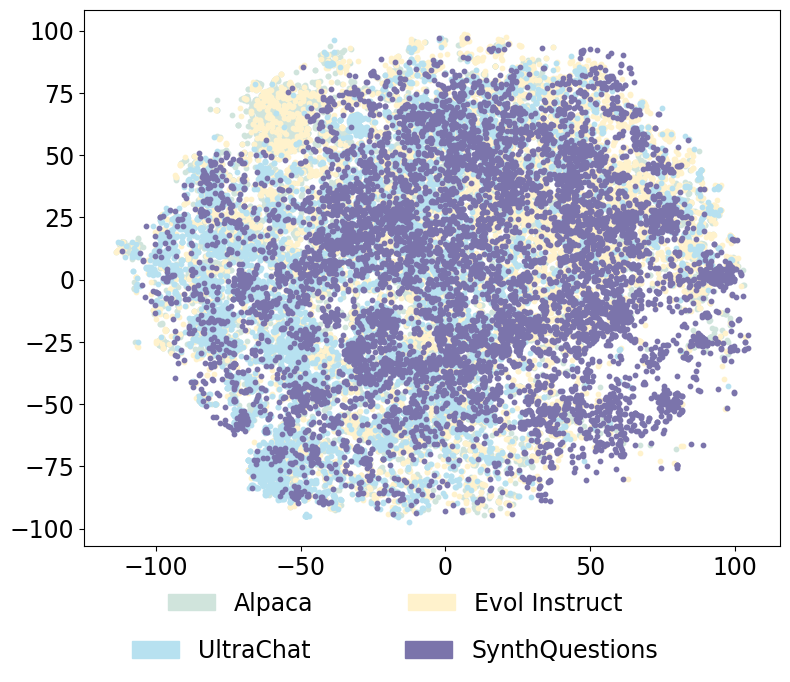}
\end{minipage}%
\begin{minipage}{.7\columnwidth}
    \centering
    \includegraphics[width=\linewidth]{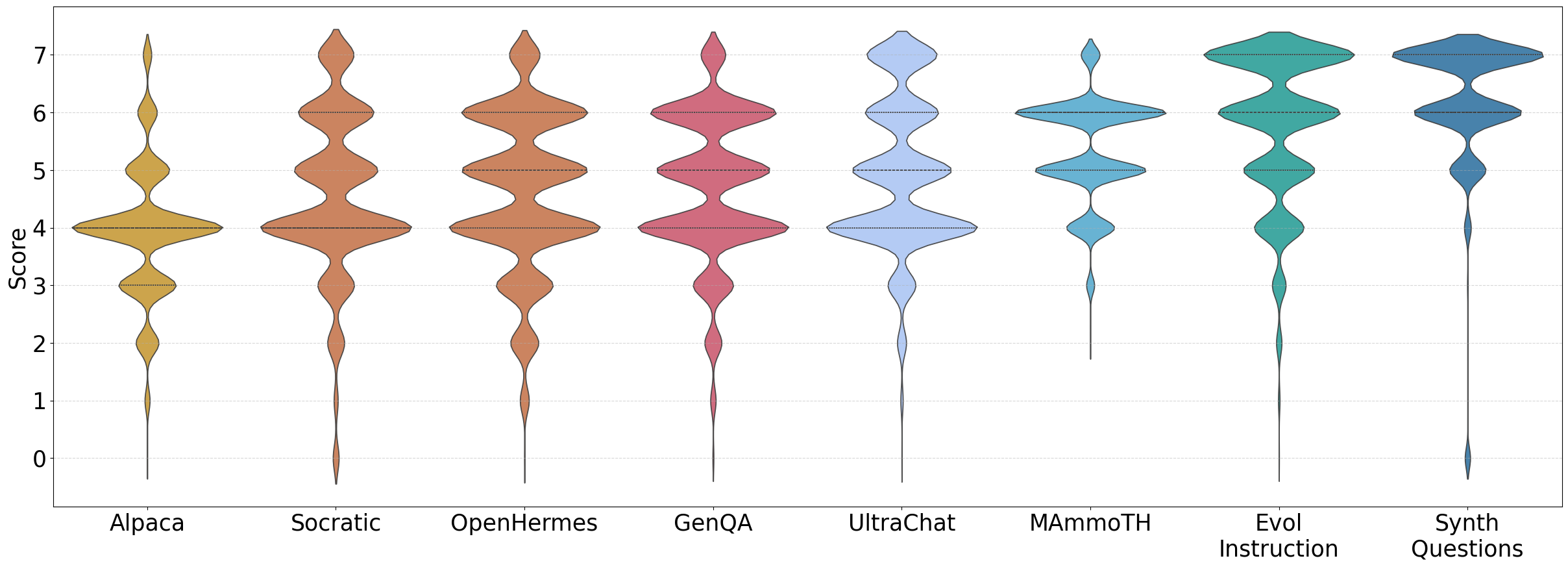}
\end{minipage}
\caption{Left: t-SNE plot of {\SynthQuestionSmallCaps} along with Alpaca, EvolInstruct and UltraChat. The t-SNE plot of {\SynthQuestionSmallCaps} covers more area than other datasets. Right: Violin plots of the Arena Scores of all synthetic datasets.}
\label{fig:visual_2}
\end{figure}

Figure~\ref{fig:visual_2} presents the full visualization of dataset diversity and complexity of different synthesized datasets.

\begin{table*}[h]
  \centering
  \begin{tabular}{ccccccccc}
    \hline
    Score &	0 &	1 &	2 &	3 &	4 &	5 &	6 &	7 \\
    \hline
    Percentage &	0.77\% &	0.0011\% &	0.021\% &	0.22\% &	2.36\% &	12.93\% &	38.83\% &	44.88\% \\
    \hline
  \end{tabular}
  \caption{The score distribution of the complete unfiltered dataset generated with our method.}
  \label{tab:full_distribution}
\end{table*}

\clearpage
\section{Dataset Safety}
\label{app:safety}
Table~\ref{tab:safety} shows the detection results from LLaMA-Guard-3-8B, which demonstrates that there are very few harmful content in our dataset.

\begin{table*}[h]
  \centering
  \resizebox{\textwidth}{!}{
  \begin{tabular}{cccccccccccccc}
    \hline
    \multirow{2}{*}{Category} &	Specialized &	Intellectual &	\multirow{2}{*}{Defamation} &	\multirow{2}{*}{Elections} &	Non-Violent &	Sexual &	Child  & \multirow{2}{*}{Privacy} &	Violent  &	\multirow{2}{*}{Hate} &	Indiscriminate  &	\multirow{2}{*}{Self-Harm} &	Sex  \\
    ~ & Advice & Property & ~ & ~ & Crimes & Content & Exploitation & ~ & Crimes & ~ & Weapons & ~ & Crimes \\
    \hline
    Percentage &	3.60\% &	0.158\% &	0.151\% &	0.098\% &	0.068\% &	0.059\% &	0.036\% &	0.034\% &	0.028\% &	0.027\% &	0.022\% &	0.013\% &	0.011\% \\
    \hline
  \end{tabular}
  }
  \caption{The percentages of detected potential harmful content in \SynthQuestionSmallCaps.}
  \label{tab:safety}
\end{table*}

\section{Full Evaluation Results}
Table~\ref{tab:chat_performance_full} and Table~\ref{tab:close_benchmarks_full} shows the full results of models on alignment benchmarks and closed-form benchmarks.
\label{app:full_eval}
\begin{table*}[h]
  \centering
  \begin{tabular}{clcccc}
    \hline
    \multirow{2}{*}{\textbf{Model Groups}} &\multirow{2}{*}{\textbf{Models}} & \textbf{Arena Hard} & \multicolumn{3}{c}{\textbf{Alpaca Eval 2.0}} \\
    & & WR(\%) & LC(\%) & WR(\%) & SD \\
    \hline
    \multirow{3}{*}{\shortstack{Larger or \\ Proprietary Models}}  & HumpBack-LLaMA2-70B & -  & 16.25 & 10.12 & 0.94\\
    & GPT-3.5-Turbo-0301 & 18.1  & 18.09 & 9.62 & 0.89      \\
    & GPT-3.5-Turbo-1106 & 18.9  & 19.30 & 9.18 & 0.89      \\
    \hline
    \multirow{3}{*}{\shortstack{7B-8B Models \\ w/ Proprietay Data}} & Mistral-7B-Instruct-v0.3 &  16.7$^{\dag}$  & 20.61 & 16.69 & 1.11 \\
    & Qwen-2-7B-Insturct$^{\dag}$ & 23.5 & 21.86 & 19.62 & 1.15    \\
    & LLaMA-3-8B-Instruct  & 20.6 & 22.92 & 22.57 & 1.26  \\
    \hline
    \multirow{10}{*}{\shortstack{LLaMA3-8B \\w/ Open-source Data}} & UltraChat & 3.6 & 8.29 & 5.44 & 0.71 \\
    & Evol Instruct & 5.1 & 8.52 & 6.25 & 0.76 \\
    & ShareGPT$^*$ & 4.8 & 9.13 & 6.55 & 0.79 \\
    & Tulu V2 Mix & 8.7 & 9.91 & 7.94 & 0.86 \\
    & OpenHermes & 4.4 & 9.94 & 6.27 & 0.73 \\
    & SocraticChat$^*$ & 10.5 & 13.85 & 9.40 & 0.87 \\
    & WildChat & 8.7 & 14.62 & 10.58 & 0.92 \\
    & GenQA & 3.0 & 9.05 & 7.11 & 0.82 \\
    & MAmmoTH2 & \textbf{16.6} & \underline{18.5} & - & -\\
    \hline
    Ours & \textbf{{\SynthQuestionSmallCaps}} & \underline{15.4} & \textbf{18.87} & \textbf{19.15} & 1.15 \\
    \hline
  \end{tabular}
  \caption{Performance of models trained on {\SynthQuestionSmallCaps} on Alpaca Eval 2.0 and Arena Hard. Among models with open-source data, the best performance is bolded and the second best performance is underlined. *marks results that are not reported and evaluated by us.}
  \label{tab:chat_performance_full}
\end{table*}

\begin{table*}[h]
  \centering
  \resizebox{\textwidth}{!}{
  \begin{tabular}{clcccccc}
    \hline
    \textbf{Model Groups} & \textbf{Models} & \textbf{IFEVAL} & \textbf{MMLU} & \textbf{ARC-C} & \textbf{GPQA} & \textbf{GSM8K} & \textbf{MATH} \\
    \hline
    \multirow{3}{*}{\shortstack{7B-8B Models \\ w/ Proprietay Data}} & Mistral-7B-Instruct-v0.3 & 54.65 & 61.84 & 63.57 & 27.8$^{\dag}$ & 43.37 & 12.94$^{\dag}$ \\
    & Qwen-2-7B-Insturct & 56.79 & 70.5 & 59.73 & 25.3 & 82.3 & 49.6 \\
    & LLaMA-3-8B-Instruct  & 74.08 & 68.5 & 82.4 & 34.6 & 80.6 & 29.1 \\
    \hline
    \multirow{8}{*}{\shortstack{LLaMA3-8B \\w/ Open-source Data}} 
    & UltraChat & - & 65.23 & 62.12 & - & 50.57 & - \\
    & Evol Instruct & - & 65.62 & 60.75 & - & 42.91 & - \\
    & ShareGPT & - & \underline{66.03} & 58.45 & - & 48.67 & - \\
    & Tulu V2 Mix & - & \textbf{66.34} & 59.22 & - & 58.07 & - \\
    & OpenHermes & - & 65.7 & 61.86 & - & 67.02 & - \\
    & WildChat & - & 65.95 & 59.22 & - & 48.75 & - \\
    & GenQA & - & 63.45 & 58.53 & - & 43.13 & - \\
    & MAmmoTH2 & 43.94$^{\dag}$ & 64.2 & \textbf{82.2} & \textbf{35.2} & \underline{70.4} & \textbf{35.8} \\
    \hline
    Ours & \textbf{{\SynthQuestionSmallCaps}} & \textbf{57.05} & 65.79 & \underline{63.92} & \underline{30.3} & \textbf{70.53} & \underline{22.71} \\
    \hline
  \end{tabular}
  }
  \caption{Performance of models on different close-ended knowledge and reasoning benchmarks. Among 7B-8B scales, the best performance is bolded and the second best performance is underlined. Results marked with \dag\; are evaluated by us.}
  \label{tab:close_benchmarks_full}
\end{table*}

\section{Results of Qwen Model}
\label{app:qwen_res}
Table~\ref{tab:qwen_res} is the results of \texttt{Qwen-2.5-7B} trained on 100K, 200K and 400K subsets of {\SynthQuestionSmallCaps}. Due to increase in base model ability, model trained on only 400K data achieves performance near to \texttt{LLaMA-3-8B} with the full 1M dataset. Model performance also shows a positive relation with dataset scale when tested on Qwen model. Due to limited compute resource, we will complete this table with 800K and 1M dataset later.

\begin{table*}[h]
  \centering
  \begin{tabular}{lccc}
    \hline
    \multirow{2}{*}{\textbf{Models}} & \multicolumn{3}{c}{\textbf{Alpaca Eval 2.0}} \\
    & LC(\%) & WR(\%) & SD \\
    \hline
    \textbf{LLaMA-3-8B}\\
    \quad{\SynthQuestionSmallCaps}-1M & 18.87 & 19.15 & 1.15 \\
    \hline
    \textbf{Qwen-2.5-7B}\\
    \quad{\SynthQuestionSmallCaps}-100K &  17.25 & 16.03 & 1.10 \\
    \quad{\SynthQuestionSmallCaps}-200K &  17.89 & 14.64 & 1.06 \\
    \quad{\SynthQuestionSmallCaps}-400K &  18.19 & 16.04 & 1.11 \\
    \hline
  \end{tabular}
  \caption{Performance of models trained on {\SynthQuestionSmallCaps} on Alpaca Eval 2.0.}
  \label{tab:qwen_res}
\end{table*}

\section{Criteria of Arena Hard}
\label{app:arena}
\begin{tcolorbox}[halign=flush left,title=Criteria from Arena Hard]
\textbf{1. Specificity}: Does the instruction ask for a specific output? \\
\textbf{2. Domain Knowledge}: Does the instruction cover one or more specific domains? \\ 
\textbf{3. Complexity}: Does the instruction have multiple levels of reasoning, components, or variables? \\
\textbf{4. Problem-Solving}: Does the instruction directly involve the AI to demonstrate active problem-solving skills? \\
\textbf{5. Creativity}: Does the instruction involve a level of creativity in approaching the problem? \\
\textbf{6. Technical Accuracy}: Does the instruction require technical accuracy in the response? \\
\textbf{7. Real-world Application}: Does the instruction relate to real-world applications?
\end{tcolorbox}

\section{Prompts and Demonstrations}
\subsection{Prompt and Demonstrations Used in Attributing Step}
\label{app:prompt}
\definecolor{babypink}{rgb}{0.96, 0.76, 0.76}
\definecolor{brilliantlavender}{rgb}{0.96, 0.73, 1.0}
\definecolor{bubblegum}{rgb}{0.99, 0.76, 0.8}
\definecolor{brightube}{rgb}{0.82, 0.62, 0.91}
\definecolor{darkpastelpurple}{rgb}{0.59, 0.44, 0.84}
\definecolor{deeplilac}{rgb}{0.6, 0.33, 0.73}
\definecolor{grayblue}{rgb}{0.54, 0.60, 0.69}
\definecolor{riceyellow}{rgb}{0.75, 0.70, 0.62}
\begin{tcolorbox}
[colback=black!5!white,colframe=teal!75!teal,title=Prompts Used in Attributing,breakable]
\lbrack \textbf{SYSTEM PROMPT}\rbrack
\begin{lstlisting}[breaklines,breakindent=0pt]
Given a document and a query to an AI assistant. 
1. You should link the document and the user query with a practical scene, considering user identity and motivation. 
2. Decompose the query regarding ability, knowledge, output and extra information: 
  - Ability: The fundamental skills or capabilities required to address the problem.
  - Knowledge: The relevant domain or subject matter related to the query.
  - Output: The expected type of response or result.
  - Extra information: Specific details or context from the scenario that ground the query in a real-world context (e.g., specific numbers, codes, or quotes from the document).

Here are some examples:
<example>
<document>
Etsy has become a leading online marketplace home to around 7.5 million active retailers, who recently generated $1.7 billion worth of revenue in a single year alone. The Etsy marketplace is excellent for selling everything from handmade creations, like home decor and digital art, to vintage products.
But is it really possible to make money on Etsy, even as a beginner? The truth is that whether or not you'll be able to make money on Etsy will largely depend on how much time you're willing to invest in learning what it takes to become a successful Etsy seller.
Fortunately, starting your own Etsy store comes with plenty of beginner-friendly benefits. The online marketplace doesn't charge mandatory monthly fees and offers plenty of great resources to help you master the easy-to-use navigation platform.
We'll walk you through everything you need to know to set up your own Etsy store and start selling in no time. You'll also get the inside scoop on what differentiates successful shops from the competition.

Main takeaways from this article:
- Setting up an Etsy store is easy - the hard part is figuring out how to make money on Etsy. We'll walk you through what you need to know to become a successful seller.
- Clothing and textiles, jewelry, personalized items, homeware, and art & collectibles are among the top-selling product categories on Etsy.
- By launching your own Etsy shop, you can sell to an established audience, minimize payment processing hassles, avoid making significant upfront investments, and adopt a multichannel selling approach.
- Working with a print on demand partner can eliminate the need to worry about purchasing supplies, keeping up with inventory, or dealing with shipping.
- Using high-quality products and providing excellent customer service are vital components that can set your shop apart.
- While setting up your Etsy shop, business licenses, taxes and fees, and shipping costs are some requirements you must take care of.
- Promoting your Etsy store through online marketing can greatly increase your odds of making money on Etsy.
</document>
<query>
Act as a an online business expert and tell me how I can use the information of the best selling products of my etsy store and use it to make more money, like listing in another website or something.
</query>
<scene>
The user might be a vendor who wants to increase the sales of his etsy store. He wants to advertise the best-selling products in his store, but has no idea where and how he can achieve this. However, he does not need suggestions that are too general without detailed and actionable guidance. He wants to seek concrete suggestions from a business expert.
</scene>
<query_compositions>
Ability: Summarizing, Planning and Guiding.
Knowledge: Business, Online store, Advertising. 
Extra Information: Etsy store
Output: A business plan or a concrete suggestion list.
</query_compositions>
</example>

<example>
<document>
Making money in stocks is usually a long-term game: Very few people make tons of money in stocks overnight. Here's how to sustainably grow your wealth with stocks.

How to make money in stocks
You can make money in stocks by opening an investing account and then buying stocks or stock-based funds, using the "buy and hold" strategy, investing in dividend-paying stocks and checking out new industries.

Open an investment account
Pick stock funds instead of individual stocks
Stay invested with the "buy and hold" strategy
Check out dividend-paying stocks
Explore new industries
</document>
<query>
You are an investment advisor, you will provide me with ideas of investments. You have $100, and your only goal is to turn that into as much money as possible in the shortest time possible, without doing anything illegal. I will do everything you say and keep you updated on our current cash total. No manual labor.
</query>
<scene>
The user might be a high-school student who wants to make some quick money to pay for his/her hobbies, but has not much principle in pocket. The fastest way to make money is without doubt investments, so he seeks investiments that do not take much principal but can earn money quickly without breaking the laws. When asking the AI assistant for suggestions, he takes $100 for an example to illustrate that he deos not has much money.
</scene>
<query_compositions>
Ability: Summarizing, Planning and Guiding.
Knowledge: Investment, Low cost investment, Business, Law. 
Extra Information: $100
Output: An investment plan or suggestions
</query_compositions>
</example>

<example>
<document>
Have you ever considered the power of a one-page website?

Modern website designs lean towards minimalism; prioritizing user experience with clean layouts, intuitive navigation, and mobile-first thinking. Less is often more!

While multi-page website architecture emphasizes structure and organization, the single-page website concept is all about simplicity and focus. It places all the vital information about your business or project on a single, scrollable page.

This can be very effective especially when you need to lead visitors to a singular action without overwhelming them with multiple pages.

In this blog post, you are going to learn how to create an effective one-page website on WordPress.com that conveys its core message and steers visitors to a specific action or understanding.

Ready to get started?
</document>
<query>
Create a one-page website for a web development company named Open Agency.
</query>
<scene>
The user might be a developer from a newly started web development company named Open Agency. The company needs a one-page website to introduce themselves, but they have not hired experts for advertising yet. As a result, the task of constructing the website is assigned to this developer. Unfortunately, he has no idea how to create such a one-page website, so he turns to an AI assistant for help with the query.
</scene>
<query_compositions>
Ability: Coding.
Knowledge: Web development, Advertising, Website creation. 
Extra Information: None
Output: A brief code snippet for a one-page website.
</query_compositions>
</example>

<example>
<document>
(some codes ...)
The error log shown is:

torch.Size([2, 12, 12])

RuntimeError                              Traceback (most recent call last)
<ipython-input-22-d2f43f09fd01> in <module>()
     74     status = 1 #F
     75     while(status == 1): #G
---> 76         qval = model(state1) #H
     77         qval_ = qval.data.numpy()
     78         if (random.random() < epsilon): #I

3 frames
/usr/local/lib/python3.7/dist-packages/torch/nn/modules/linear.py in forward(self, input)
    101 
    102     def forward(self, input: Tensor) -> Tensor:
--> 103         return F.linear(input, self.weight, self.bias)
    104 
    105     def extra_repr(self) -> str:

RuntimeError: mat1 and mat2 shapes cannot be multiplied (128x4 and 128x64)
mat1 should be the output of the convolutional network after it is flattened, and mat2 is the linear network following it. Appreciate any help. Thanks!
</document>
<query>
I'm initializing my observation as np.zeros((111,))  and state representation is as follows: 109 Laser scan points, yaw and  distance to goal total 111. I don't know why I'm getting the following error: [ERROR] [1684308219.676930, 2100.420000]: bad callback: <bound method EvaderNode.scan_callback of <__main__.EvaderNode object at 0x7f77a26aaca0>>
Traceback (most recent call last):
  File "/opt/ros/noetic/lib/python3/dist-packages/rospy/topics.py", line 750, in _invoke_callback
    cb(msg)
  File "/home/cse4568/catkin_ws/src/pa2/src/evader_2.py", line 636, in scan_callback
    self.agent.train(32) # Set the batch size here
  File "/home/cse4568/catkin_ws/src/pa2/src/DQN.py", line 64, in train
    target = reward + self.gamma * torch.max(self.q_target(torch.tensor([next_state], dtype=torch.float32)))
  File "/home/cse4568/.local/lib/python3.8/site-packages/torch/nn/modules/module.py", line 1110, in _call_impl
    return forward_call(*input, **kwargs)
  File "/home/cse4568/catkin_ws/src/pa2/src/DQN.py", line 27, in forward
    return self.model(x)
  File "/home/cse4568/.local/lib/python3.8/site-packages/torch/nn/modules/module.py", line 1110, in _call_impl
    return forward_call(*input, **kwargs)
  File "/home/cse4568/.local/lib/python3.8/site-packages/torch/nn/modules/container.py", line 141, in forward
    input = module(input)
  File "/home/cse4568/.local/lib/python3.8/site-packages/torch/nn/modules/module.py", line 1110, in _call_impl
    return forward_call(*input, **kwargs)
  File "/home/cse4568/.local/lib/python3.8/site-packages/torch/nn/modules/linear.py", line 103, in forward
    return F.linear(input, self.weight, self.bias)
RuntimeError: mat1 and mat2 shapes cannot be multiplied (1x113 and 111x128)
And everytime it runs I'm getting different mat1 values. Find where I made the mistake and fix the code. You are welcome to make all the necessary changes and modfications to make it the best DQN implementation for my Autonomous robot navigation in maze like env. I already implemented the Evader node. You can modify the DQN to make it fit for the Evader:
(some codes ...)
</query>
<scene>
The user might be a student studying reinforcement learning, who is developing an algorithm based on DQN model. However, he is faced with an error "mat1 and mat2 shapes cannot be multiplied" in his code. He is not familiar with pytorch, so he copied his error log and codes to ask the assistant to debug for him.
</scene>
<query_compositions>
Ability: Coding, Debugging.
Knowledge: Python, PyTorch, Deep Learning. 
Extra Information: A code snippet copied from the document (Traceback...).
Output: The corrected code or suggestions on how to fix the bug.
</query_compositions>
</example>
\end{lstlisting}
\lbrack \textbf{USER PROMPT}\rbrack
\begin{lstlisting}[breaklines,breakindent=0pt]
Now imagine a practical scene which link the user query and the document. Describe such a scene with one brief paragraph, containing the user identity and the motivation. Then also decompose the query regarding ability, knowledge, extra information and output.

Remember you are not responding the query. Only output with the following JSON format without any additional explanation or chat:
{{
    "scene": "xxx",
    "query_compositions": {{
        "ability": "xxx",
        "knowledge": "xxx",
        "extra_information": "xxx",
        "output": "xxx"
    }}
}}

## Document
{document}

## Query
{query}

## Scene
\end{lstlisting}
\end{tcolorbox}

\subsection{Prompts Used in Synthesizing Step}
\begin{tcolorbox}[colback=black!5!white,colframe=olive!75!olive,title=Prompts Used in Synthesizing, breakable]
\lbrack \textbf{SYSTEM PROMPT}\rbrack
\begin{lstlisting}[breaklines,breakindent=0pt]
You will be shown a document, you should imagine a scene where a user with a certain identity comes up with some query compositions and a query related to the document. Here are some examples:

{demos}
\end{lstlisting}

\lbrack \textbf{USER PROMPT}\rbrack
\begin{lstlisting}[breaklines,breakindent=0pt]
Now you should
1. Envision a real-world scenario based on the provided document. Describe this scenario in one paragraph, detailing the logical steps from the document's content to a query directed at an AI assistant.
2. Then list the compositions of a query that could emerge from this scenario, including:
    - Ability: The fundamental skills or capabilities required to address the problem.
    - Knowledge: The relevant domain or subject matter related to the query.
    - Output: The expected type of response or result.
    - Extra information: Specific details or context from the scenario that ground the query in a real-world context (e.g., specific numbers, codes, or quotes from the document).
3. Finally formulate a user query based on the scenario and query compositions you have identified. Ensure:
    - Maximize the ability that is needed to solve the query. Avoid simple copying or extracting tasks.
    - The query should be practical, complex and requires advanced skills. It should be challenging for the most capable AI.
    - The query should be self-contained and answerable without additional resources.
    - You must copy exerpts from the document into the query if extra information from the document is needed.
    - As the AI assistant does not have search engine access, **avoid** creating queries that rely on external search engines.

When constructing query compositions and the final query, consider the following requirements:
> Specificity: The query should ask for a specific output;
> Domain Knowledge: The query should cover one or more specific domains;
> Complexity: The query should have multiple levels of reasoning, compositions, or variables;
> Problem-Solving: The query should directly involve the AI to demonstrate active problem-solving skills;
> Creativity: The query should involve a level of creativity in approaching the problem;
> Technical Accuracy: The query should require technical accuracy in the response;
> Real-world Application: The query should relate to real-world applications.
    
Output the scene and query in JSON format. Before generating scene, query_composition and query, you should include your thought on how you design the real-world scenario and the query, so that each of the above requirements is satisfied.

## Document
{document}

## Output Format
{{
    "thought": "xxx"
    "scene": "xxx",
    "query_compositions": {{
        "ability": "xxx",
        "knowledge": "xxx",
        "extra_information": "xxx",
        "output": "xxx"
    }},
    "query": "xxx"
}}

## Your Output
\end{lstlisting}
\end{tcolorbox}

\subsection{Prompts Used for Filtering Instructions}
\begin{tcolorbox}[colback=black!5!white,colframe=black!75!black,title=Prompts Used for Filtering Instructions, breakable]
\begin{lstlisting}[breaklines,breakindent=0pt]
## Role
Prompt Evaluator

## Task
You will be given a prompt written for large language models, and you should evaluate the prompt accoring to the provided criteria.

## Evaluation Criteria
1. Specificity: Does the prompt ask for a specific output?
2. Domain Knowledge: Does the prompt cover one or more specific domains?
3. Complexity: Does the prompt have multiple levels of reasoning, compositions, or variables?
4. Problem-Solving: Does the prompt directly involve the AI to demonstrate active problem-solving skills?
5. Creativity: Does the prompt involve a level of creativity in approaching the problem?
6. Technical Accuracy: Does the prompt require technical accuracy in the response?
7. Real-world Application: Does the prompt relate to real-world applications?

## Rules
1. You should evaluate based on each aspects of the criteria independently. First analyze the prompt according to each aspect and then assign it with a score.
2. If a prompt satisfies one aspect, you should score it as 1. Otherwise you should score it as 0.
3. Output your results with JSON dictionary format. 

## Output Sample
{
    "specificity": {"analysis": "analysis about specificity", "score": n},
    "domain_knowledge": {"analysis": "analysis about domain knowledge", "score": n},
    "complexity": {"analysis": "analysis about complexity", "score": n},
    "problem_solving": {"analysis": "analysis about problem solving", "score": n},
    "creativity": {"analysis": "analysis about creativity", "score": n},
    "technical_accuracy": {"analysis": "analysis about technical accuracy", "score": n},
    "real_world_application": {"analysis": "analysis about real-world application", "score": n}
}
Here is the prompt to evaluate:
{prompt}
\end{lstlisting}
\end{tcolorbox}

\section{Other Cases}
\label{app:cases}
\begin{tcolorbox}[colback=black!5!white,colframe=darkpastelpurple!75!darkpastelpurple,title=Case 1, breakable]
\lbrack \textbf{Document}\rbrack \\
You hear a lot about how creatives and account people don't get along, that couldn't be farther from the truth at this place. We eat lunch together, we make fun of each other and we laugh together. Soon we will be playing ping pong together. We are one big family. To my office family, thank you for making me feel so welcome in such a short amount of time. I can't wait to see what 2013 has in store for us. \\\\
\lbrack \textbf{User}\rbrack \\
A new employee in a creative agency. \\\\
\lbrack \textbf{Motivation}\rbrack \\
The user is trying to understand the company culture and dynamics. They want to know how to foster and maintain a positive and collaborative atmosphere in their own team. \\\\
\lbrack \textbf{Instruction}\rbrack \\
What are some actionable steps to foster a positive and collaborative team culture, like being 'one big family', in a creative agency, and how can we maintain it over time?
\end{tcolorbox}

\begin{tcolorbox}[colback=black!5!white,colframe=darkpastelpurple!75!darkpastelpurple,title=Case 2, breakable]
\lbrack \textbf{Document}\rbrack \\
Your health your way Kids may not think to pay much attention to their health. As long as they can move around when they want and have a regular source of food their health and well being takes back seat to other more entertaining thoughts.But with their peers increasingly becoming overweight and obese those concerns are starting to be pushed to the front of the line whether kids want to address the problem or not. Living healthy not only decreases chance of health risks but can also make a child's live much more well rounded and enjoyable. Living healthy isn't hard, it's all about making the right choices and creating good habits. In order to kick a healthy lifestyle into high gear kids have to get away from their video games and televisions and get moving.\\\\
\lbrack \textbf{User}\rbrack \\
A health-conscious parent. \\\\
\lbrack \textbf{Motivation}\rbrack \\
The user is browsing the internet for ways to encourage their child to adopt a healthy lifestyle. They come across an article emphasizing the importance of living healthy, especially for kids, and want to create a personalized plan for their child to make healthy choices and develop good habits. \\\\
\lbrack \textbf{Instruction}\rbrack \\
Design a personalized health plan for a 10-year-old child who spends most of their time playing video games and has a sedentary lifestyle. The plan should include a daily routine, dietary recommendations, and fun physical activities that can help them develop healthy habits and reduce the risk of obesity. Consider the child's age, interests, and current lifestyle when creating the plan. Provide a step-by-step guide on how to implement the plan and make adjustments as needed.
\end{tcolorbox}

\begin{tcolorbox}[colback=black!5!white,colframe=darkpastelpurple!75!darkpastelpurple,title=Case 3, breakable]
\lbrack \textbf{Document}\rbrack \\
One of the many challenges that face senior citizens is keeping their home clean. A clean house can do wonders for your mental health and well-being, and no one should have to sacrifice the joy of a clean and presentable home because chores have become too difficult to complete. Chrisel’s Affordable Care is your choice for in-home care and cleaning services in the Suffield, CT area. Often, keeping up with a loved one’s house cleaning is left up to the family. Our job is to take the stress off you or your loved one’s plate, so let us take care of those responsibilities for you. Chrisel’s offers house cleaning packages that fit every budget. Contact us today to find out how we can help you. You don’t want to let just anyone into your home. You trust Chrisel’s Affordable Care with your loved ones, so why not trust us to keep their home neat and tidy? We’re insured and bonded, and we offer free consultations. Every family has different needs. Chrisel’s Affordable Care offers a basic cleaning service that covers your most immediate cleaning ones. This basic package includes:\\\\
\lbrack \textbf{User}\rbrack \\
A family caregiver. \\\\
\lbrack \textbf{Motivation}\rbrack \\
The user is exploring in-home care and cleaning services for their elderly loved one in the Suffield, CT area. They come across Chrisel's Affordable Care and want to determine the best cleaning package for their loved one's needs and budget. \\\\
\lbrack \textbf{Instruction}\rbrack \\
Considering the basic cleaning service package offered by Chrisel's Affordable Care, how can I create a customized cleaning package that meets my elderly loved one's specific needs and stays within their budget? Please provide a cost-benefit analysis of the recommended package and explain why it's the most suitable option.
\end{tcolorbox}

\begin{tcolorbox}[colback=black!5!white,colframe=darkpastelpurple!75!darkpastelpurple,title=Case 4, breakable]
\lbrack \textbf{Document}\rbrack \\
<p>The easiest way is to create a storyboard that change opacity of two images with blend. Then you can call the storyboard.begin() to run the storyboard you want for example from the tapped or click event.</p>\\\\
\lbrack \textbf{User}\rbrack \\
A mobile app developer. \\\\
\lbrack \textbf{Motivation}\rbrack \\
The user is tasked with creating an engaging user interface for a new app. They want to implement an animation that transitions between two images with a blend effect. The developer has experience with storyboard animations but needs help with implementing this specific effect. \\\\
\lbrack \textbf{Instruction}\rbrack \\
Create a storyboard animation that transitions between two images with a blend effect, where the opacity of the images changes, and provide a code snippet or step-by-step guide on how to implement it in a mobile app.
\end{tcolorbox}

\end{document}